\documentclass[journal]{IEEEtran}
\usepackage{amsmath,amsfonts}
\usepackage{algorithmic}
\usepackage{array}
\usepackage{textcomp}
\usepackage{stfloats}
\usepackage{url}
\usepackage{verbatim}
\usepackage{graphicx}
\usepackage{amssymb}
\usepackage{cite}
\usepackage{hyperref}
\usepackage{booktabs}
\usepackage{threeparttable}
\usepackage{bm}
\usepackage{color}
\usepackage{arydshln}
\usepackage{microtype}
\usepackage{makecell}
\usepackage{multirow}
\usepackage{times}
\usepackage{latexsym}
\usepackage{float}
\usepackage{subfigure}
\usepackage{tabularx}
\usepackage{xcolor}
\usepackage{stmaryrd}

\usepackage{blindtext}
\usepackage{tablefootnote}
\def\BibTeX{{\rm B\kern-.05em{\sc i\kern-.025em b}\kern-.08em
    T\kern-.1667em\lower.7ex\hbox{E}\kern-.125emX}}
\usepackage{balance}
\begin{document}
\begin{sloppypar}
\title{Pathformer: Recursive Path Query Encoding for Complex Logical Query Answering}
\author{Chongzhi Zhang,
    ~Zhiping~Peng,
    ~Junhao~Zheng,
        ~Linghao~Wang,
        ~Ruifeng~Shi,
    Qianli~Ma*,~\IEEEmembership{Member,~IEEE}
    \thanks{Chongzhi Zhang, Junhao Zheng, Linghao Wang, Ruifeng Shi, and Qianli Ma are with the School of Computer Science and Engineering, South China University of Technology, Guangzhou, 510006, China (e-mail:  cschongzhizhang@mail.scut.edu.cn, qianlima@scut.edu.cn, *corresponding author). }
    \thanks{Zhiping Peng is with Guangdong University of Petrochemical Technology, Maoming, 525000, China, and Jiangmen Polytechnic, Jiangmen, 529000, China. (e-mail: pengzp@foxmail.com)}
    }

\markboth{Journal of \LaTeX\ Class Files,~Vol.~18, No.~9, September~2020}%
{Pathformer: Recursive Path Query Encoding for Complex Logical Query Answering}

\maketitle

\begin{abstract}
Complex Logical Query Answering (CLQA) over incomplete knowledge graphs is a challenging task. Recently, Query Embedding (QE) methods are proposed to solve CLQA by performing multi-hop logical reasoning. However, most of them only consider historical query context information while ignoring future information, which leads to their failure to capture the complex dependencies behind the elements of a query. In recent years, the transformer architecture has shown a strong ability to model long-range dependencies between words. The bidirectional attention mechanism proposed by the transformer can solve the limitation of these QE methods regarding query context. Still, as a sequence model, it is difficult for the transformer to model complex logical queries with branch structure computation graphs directly. To this end, we propose a neural one-point embedding method called Pathformer based on the tree-like computation graph, i.e., query computation tree. Specifically, Pathformer decomposes the query computation tree into path query sequences by branches and then uses the transformer encoder to recursively encode these path query sequences to obtain the final query embedding. This allows Pathformer to fully utilize future context information to explicitly model the complex interactions between various parts of the path query. Experimental results show that Pathformer outperforms existing competitive neural QE methods, and we found that Pathformer has the potential to be applied to non-one-point embedding space. 
\end{abstract}


\section{Introduction} 
\label{Introduction}
\IEEEPARstart{K}{nowledge} Graphs (KGs) have drawn much attention in academia and industry, 
which have been often used in various intelligent application scenarios \cite{wang2023mixed, xiong2017explicit, atif2023beamqa, yang2022knowledge}. 
However, due to the limitations of KG construction techniques \cite{toutanova2015observed, vrandevcic2014wikidata}, real-world KGs \cite{bollacker2008freebase, suchanek2007yago, carlson2010toward} are often considered noisy and incomplete, which is also known as the Open World Assumption (OWA) \cite{libkin2009open}. 
Complex Logical Query Answering (CLQA) is a fundamental yet challenging task which requires answering the Existential First Order Logic (EFOL) queries with logical operators, including existential quantifier ($\exists$), conjunction ($\wedge$), disjunction ($\vee$), and negation ($\neg$) on incomplete KGs. 
Since CLQA is geared toward KGs with OWA, 
the method to solve CLQA needs to perform multi-hop logical reasoning over KGs. 
That is, not only to execute logical operators but also to utilize available knowledge to predict the unseen one, which is the root cause of the failure of traditional traversal methods to answer complex logical queries \cite{ren2020query2box}. 

\begin{figure*}
    \centering
    \includegraphics[width=\textwidth]
    {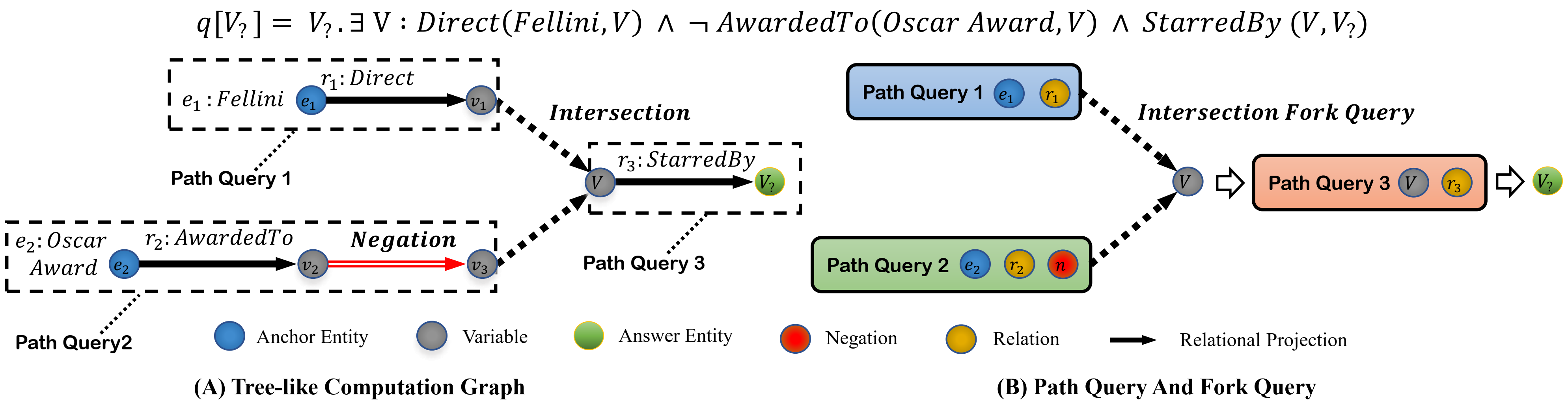}
    \caption{
    (A): Given a query \emph{"Who starred in Fellini's films that didn't win any Oscar awards?"}, we can get the corresponding logical formula and tree-like computation graph. 
    The branch path sequences in dashed boxes are essentially path queries. (B): Pathformer decomposes a tree-like EFOL query into path queries and fork queries. After getting the representation of the intermediate existential variable $V$ by encoding the fork query, we can continue encoding the $\mathrm {Path\ Query\ 3}$ to obtain the representation of the single free variable (i.e., answer entity) $V_{?}$. }
    \label{figure1}
\end{figure*}

Recently, Query Embedding (QE) methods \cite{hamilton2018embedding, ren2020query2box, ren2020beta, amayuelas2022neural, zhang2021cone, bai2022query2particles, yang2022gammae, zhang2024conditional} have been proposed to solve CLQA. 
QE methods embed entities and queries into continuous vector spaces to select answer entities by comparing the distance between the candidate entity embeddings and query embedding. 
They represent the EFOL query formula as a computation graph, where logic operations are replaced with corresponding set operations. 
Moreover, the existential quantifier induces a set projection operation, which corresponds to logic skolemization \cite{luus2021logic}. 
For example, a query \emph{"Who starred in Fellini's films that didn't win any Oscar awards?"} can be converted into an EFOL query with a single free variable: $V_{?} . \exists  V : Direct(Fellini, V) \wedge  \neg AwardedTo(Oscar\ Award, V) \wedge StarredBy(V, V_{?})$. 
As shown in Figure \ref{figure1} (A), we can obtain a tree-like computation graph based on the query. 
Each node in the computation graph represents an entity or an entity set, while each edge represents the corresponding set logic operation. 

However, although most QE methods effectively tackle the incompleteness of KG, there is still a limitation. 
They usually encode the query link by link starting from the anchor entity nodes according to the computation graph. 
These approaches only consider previous query context information but not future context, 
making it difficult to model the implicit complex dependencies between various parts of a query. 
BIQE \cite{kotnis2021answering} is also aware of this point, so it uses a transformer encoder \cite{devlin2018bert} to model the entire computation graph, thereby fully utilizing the query context information through the bidirectional attention mechanism \cite{vaswani2017attention}. 
To represent the structure of the computation graph, BIQE designs a special positional encoding scheme. 
However, according to current studies on graph transformers \cite{ma2023GraphInductiveBiases, ying2021transformers}, 
graph inductive biases are crucial for the transformers to encode graphs, 
and only introducing positional encoding may be insufficient for modelling graph structure information. 
This means the query encoding approach of BIQE may lose the structural information of the input query graph. 
Additionally, BIQE cannot handle negative queries. 

In this paper, we focus on an important subset of EFOL queries whose computation graphs are trees, i.e., query computation trees. 
For a tree, all nodes except leaf ones are branch nodes. 
As is shown in the tree-like computation graph in Figure \ref{figure1} (A), the anchor entity nodes are leaf nodes, and others are branch nodes. 
For example, the existential variable node $V$ is a two-branch node, and each branch can be regarded as a path query sequence. 
Similarly, the free variable node (i.e., answer node) $V_{?}$  is a single-branch node, and its branch $V\ \stackrel{r_{StarredBy}}{\longrightarrow}\ V_{?}$ is also a path query sequence. 
This means that we can decompose a tree-like query into path query sequences by branches,  which allows us to follow the query computation tree to solve the query path by path. 
For these path query sequences, we can naturally introduce the transformer \cite{vaswani2017attention} to encode them separately. 
Through the bidirectional attention mechanism, we can fully utilize future context to explicitly model the complex dependencies between various parts of a path query. 
In contrast to BIQE, the transformer is not used to encode the entire query computation tree but a branch path query sequence, 
so there is no need to consider the loss of input graph structure information caused by insufficient graph inductive biases for the transformer. 

Therefore, we propose a neural QE method named Pathformer, which decomposes a tree-like query into path query sequences by branches. 
We define the fork query as the process of aggregating the representations of each branch path query sequence to obtain the corresponding multi-branch variable node representation, 
so a tree-like EFOL query can be viewed as a combination of path queries and fork queries, as shown in Figure \ref{figure1} (B). 
We use the encoder part of the transformer \cite{vaswani2017attention} as the path query encoder to encode the entire path query sequence at once. 
Furthermore, by introducing just one token representing the set complement/negation operation, 
Pathformer can encode path queries with negation. 
By encoding the fork query, 
we can get the representations of the intermediate multi-branch variable nodes to complete the subsequent path queries. 
Thus, we can recursively encode all the path query sequences through the fork queries to embed the complex query into the one-point vector space. 
We conducted experiments on two KG reasoning benchmark datasets: FB15k-237 \cite{toutanova2015observed} and NELL995 \cite{xiong2017deeppath}. 
The experimental results show that Pathformer can outperform existing competitive neural QE methods, 
and we find that Pathformer, as a one-point embedding method, has the potential to be applied to non-one-point embedding spaces \cite{ren2020query2box, ren2020beta} after experimental analyses.

\section{Related Work}
\label{Related Work}

\paragraph{Knowledge Graph Reasoning}

Traditional KG reasoning tasks such as KG completion \cite{socher2013reasoning} are essentially one-hop query problems. 
That is, only one link is involved. 
Representative methods on KG completion are Knowledge Graph Embedding (KGE) methods \cite{bordes2013translating, sun2019rotate, trouillon2016complex, chen2022meta, zhang2023weighted}, 
which embed entities and relations into vector spaces and make inferences by scoring triples with a well-defined scoring function. 
Other methods for KG completion include rule-learing \cite{sadeghian2019drum, zhang2019iteratively}, text representation learning \cite{wang2021kepler, saxena2022sequence, wang2021structure}, and GNNs \cite{zhang2022rethinking, zhang2022knowledge}. 
However, the CLQA task is a multi-hop query problem since finding the answers to a complex logical query on a large and incomplete KG may involve multiple variables and unobserved links. 
Moreover, compared to answering one-hop queries on incomplete KGs, CLQA also requires the ability to execute logical operators. 
Therefore, CLQA can be regarded as a multi-hop logical reasoning task on KGs. 

\paragraph{Neural Complex Logical Query Answering}

Recently, some neural QE methods have been proposed to solve the CLQA problem. 
They embed EFOL queries and entities into the vector space and select answer entities based on the distance of query embeddings to candidate entity embeddings. 
Motivated by the advancements in deep learning on sets \cite{zaheer2017deep}, GQE \cite{hamilton2018embedding} embeds queries and entities into a one-point vector space to support conjunctive queries. 
Query2box \cite{ren2020query2box} designs various set logic operations based on box embedding and uses Disjunctive Normal Form (DNF) to convert disjunctive queries into conjunctive queries to support the union operation corresponding to disjunction in a scalable manner. 
BetaE \cite{ren2020beta} models entities and queries with beta probability distributions to support the negation operator. 
ConE \cite{zhang2021cone} is also a geometric embedding method, which enables the negation operator through cone embedding. 
FuzzQE \cite{chen2022fuzzy} uses t-norm fuzzy logic \cite{klement2013triangular} to improve previous embedding methods. 
Query2Particles \cite{bai2022query2particles} uses the transformer \cite{vaswani2017attention} to encode each query into multiple particle vectors based on the diversity of answers. 
MLP\&MLP-Mixer \cite{amayuelas2022neural} uses a set of neural-network-based models to embed entities and queries into a one-point vector space. 
GammaE \cite{yang2022gammae} uses gamma embeddings to encode complex logical queries. 
BIQE \cite{kotnis2021answering} uses a bidirectional sequence encoder \cite{devlin2018bert} to solve conjunctive queries. 
Our work shares a similar spirit with BIQE in the sense that both methods use the transformer encoder to learn future information to model the implicit complex dependencies between various parts of a query.
However, BIQE processes the entire computation graph at a time, and for this purpose, 
it designs particular positional encoding scheme to represent graph structure and conjunction operator.  
As a result, it cannot supports negation operator. 
In addition, according to the studies on graph inductive biases \cite{ma2023GraphInductiveBiases, ying2021transformers}, it is insufficient to represent the graph structure only by positional encoding scheme. 
This insufficiency may result in the loss of input information, thereby affecting performance. 
Pathformer just uses the transformer encoder to recursively encode path query sequences without the need to represent the graph structure. 
Furthermore, Pathformer can handle negative queries. 

In addition to neural QE methods, some recent studies incorporate symbolic information into neural methods. We discuss these symbolic integration methods in Supplementary Material \ref{ns method}.

\section{Preliminaries}
\label{Preliminaries}

A Knowledge Graph (KG) $\mathcal{G}$ consists of a set of entities $\mathcal{V}$ and a set of relation types $\mathcal{R}$. It can be defined as a set of triples $\mathcal{E} ={(h_{i},r_{i},t_{i})} \subseteq \mathcal{V} \times \mathcal{R} \times \mathcal{V}$, 
where each triple indicates there is a directed edge of relation type $r_{i}$ from head entity $h_{i}$ to tail entity $t_{i}$, 
namely $\mathcal{G}=(\mathcal{V}, \mathcal{E}, \mathcal{R})$. 
A KG can be represented as a first-order logic knowledge base, 
where each triple $(h_{i},r_{i},t_{i})$ can be viewed as an atomic formula $r_{i}(h_{i},t_{i})$, with relation type $r_{i}: \mathcal{V} \times \mathcal{V}\to \left \{ True,False \right \}$ is a binary predicate \cite{ren2020beta, arakelyan2020complex}.

\paragraph{Existential First Order Logic Queries}
We focus on answering complex logical queries on incomplete KGs. 
Similar to previous works \cite{hamilton2018embedding, ren2020beta, amayuelas2022neural}, 
we consider Existential First Order Logic (EFOL) queries with only one free variable. 
Such queries are a subset of first-order logic queries, using existential quantification ($\exists$), conjunction ($\wedge$), disjunction ($\vee$), and atomic negation ($\neg$). 
Following the definition in \cite{ren2020beta}, such queries can be expressed with disjunctive normal form \cite{davey2002introduction} as follows: 
\begin{displaymath}
\begin{split}
    q[V_{?}] = V_{?}. \exists V_{1},...,V_{k}: (a_{1}^{1}\wedge ...\wedge a_{n_{1}}^{1})\vee... \vee (a_{1}^{d}\wedge ...\wedge a_{n_{d}}^{d}),
\end{split}
\end{displaymath}
where $V_{?}$ is the only free variable, $ V_{1},...,V_{k}$ are $k$ existential variables. 
$a_{j}^{i}$ are atomic formulas with constant entities and variables. They can be either negated or not: 
\begin{displaymath}
\begin{split}
     a_{j}^{i}=\begin{array}{l} 
  \left\{\begin{matrix} 
  r(e_{a},V)\ or\ r(V,V') \\ 
  \neg r(e_{a},V)\ or\ \neg r(V,V') 
\end{matrix}\right.    
\end{array}
,
\end{split}
\end{displaymath}
where $e_{a} \in \mathcal{V}$ is a input constant (anchor) entity, $V,V'\in \{V_{?},V_{1},...,V_{k}\}$ are variables, $V\ne V'$.

\paragraph{Query Computation Tree and Set Logic Operators}

Following the previous studies \cite{amayuelas2022neural,ren2020query2box,ren2020beta,hamilton2018embedding}, we also consider using the computation graph to convert first-order logic operators to set logic operators to answer logical queries. 
In particular, the existential quantifier induces a set relational projection operation, which corresponds to logic skolemization \cite{luus2021logic}. 
In this paper, we focus on the tree-like query,
which is more in line with the human multi-hop query style \cite{bai2023answering}. 
As shown in Figure \ref{figure1} (A), given a query, we can get its corresponding tree-like computation graph (i.e., query computation tree) by representing each atomic formula with relational projection, merging by intersection and transforming negation by complement. The tree shows 
the computation process of how to answer the query. 
Each leaf node in the query computation tree corresponds to the input anchor entity $e_{a} \in \mathcal{V}$, each branch node is an intermediate variable, and in particular, the root node represents the single free variable of the query, which is a set of answer entities. 
The mapping along each directed edge applies a certain set logic operator: 
\begin{itemize}
\item \emph{Relational Projection:} Given a set of entities $\mathcal{S} \subseteq \mathcal{V}$  and a relation $r\in \mathcal{R}$, 
the projection operator returns a new entity set $\mathcal{S}'$, each entity in $\mathcal{S}'$ is connected to one or more entities in $\mathcal{S}$ by an edge of relation type $r$, i.e. $\mathcal{S}'=\{e'\in \mathcal{V}\ |\ r(e,e')=True, e\in \mathcal{S}\}$. 
\item \emph{Intersection:} Given several sets of entities $\{ \mathcal{S}_{1},\mathcal{S}_{2},...,\mathcal{S}_{n}\}$, the intersection operator returns their intersection ${\textstyle \bigcap_{i=1}^{n}\mathcal{S}_{i}}$. 
\item \emph{Complement/Negation:} Given a set of entities $\mathcal{S}$, calculate its complement $\bar{\mathcal{S}} = \mathcal{V}- \mathcal{S}$. 
\end{itemize}
There is no need to define a set union operator corresponding to disjunction, as demonstrated by Query2box \cite{ren2020query2box}. 
It illustrates that disjunctive queries can be handled in a scalable manner by transforming queries into a Disjunctive Normal Form (DNF) \cite{davey2002introduction}. 

To address the complex logical query $q$, the corresponding answer entity set $\llbracket q \rrbracket \subseteq \mathcal{V}$ needs to be determined, 
where $\llbracket q \rrbracket$ is a set of entities such that $e \in \llbracket q \rrbracket$ iff $q[e]=True$. 
In order to obtain such an answer entity set $\llbracket q \rrbracket$, 
we can follow the query computation tree, execute set logic operators, 
embed the query into vector space, 
and derive $\llbracket q \rrbracket$ from the entities that are close to the query in the embedding space. 
Although such traversal of query computation tree is similar to traversing the KG, traditional KG traversal cannot handle missing edges in the incomplete KG \cite{ren2020beta}.

\begin{figure*}
    \centering
    \includegraphics[width=\textwidth]{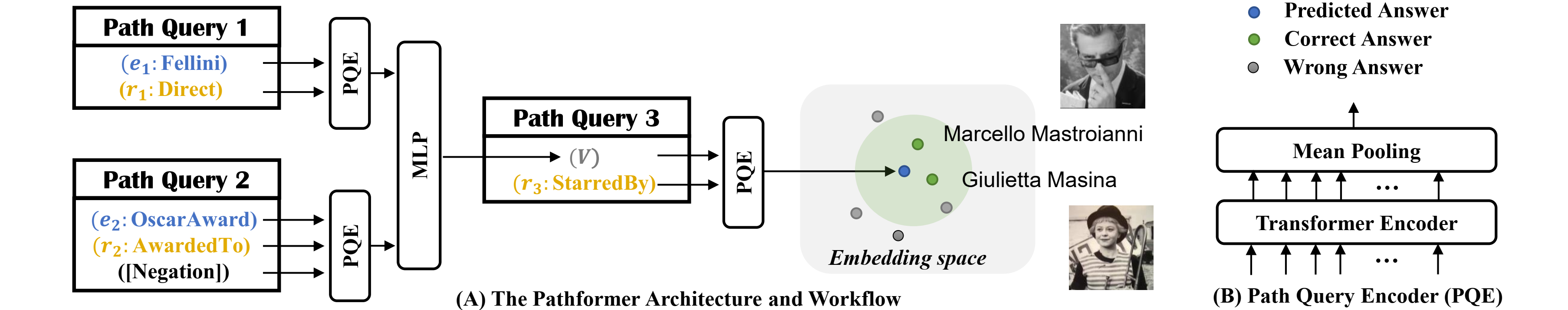}
    \caption{The Pathformer architecture. ($\star $) denotes the embedding of $\star $. 
    Pathformer can recursively encode path queries to obtain the final query embedding. 
    After obtaining the embedding of the multi-branch variable node $V$ by aggregating the embeddings of each branch path query, we can continue to encode $\mathrm {Path\ Query\ 3}$ to get the predicted answer embedding, then perform distance-based retrieval to find the correct answer in the embedding space. }
    \label{figure2}
\end{figure*}

\section{Proposed Method}
\label{others}

In this section, we first introduce how to decompose a complex logical query into path queries and fork queries. 
Then, we define a path query encoder and show how to get the input sequence of the encoder from a path query. 
Accordingly, we also define a fork query encoder for fork queries. 
Finally, we represent the training objective of our method.

\subsection{Query Computation Tree Decomposition}
As shown in Figure \ref{figure1} (A), a complex logical query can be represented as a tree-like computation graph, which we refer to as a query computation tree. 
Except for the leaf nodes of a tree, other nodes, including the root one, are all branch nodes, and each branch node has its branch(es). 
This means we can decompose a query computation tree into multiple path sequences by branches. 
We use the anchor nodes as the starting nodes to traverse the query computation tree to obtain the path sequence recursively. 
The starting node refers to the node from which the current traversal process starts. 
When traversing to the multi-branch node, we can get all the branch path sequences from the starting nodes to the child nodes of the multi-branch node and take the multi-branch node as the new starting node if the multi-branch node is not the root node. 
When traversing to the root node, we can obtain the branch path sequence from the starting node to the root node. 

For example, as illustrated in Figure \ref{figure1}, 
we can obtain three paths which essentially are path queries. 
For $\mathrm {Path\ Query\ 2}$, its starting node is anchor entity node $e_{2}$, and it has a set operator sequence: $\{r_{2}\ projection, negation\}$. Similarly, For $\mathrm {Path\ Query\ 3}$, the starting node and set operator sequence are variable node $V$ and $\{r_{3}\ projection\}$, respectively. 
Since a path query encoded by our method is based on the starting node and set operator sequence of the path (it will be later discussed in Sections \ref{Path Query Encoder}), the path sequence obtained by taking the known anchor node as the starting node is a complete path query and can be encoded independently. 
However, the path sequence obtained from a variable node as the starting node cannot be independently encoded because the variable node is unknown, such as $\mathrm {Path\ Query\ 3}$. 
Obviously, the variable nodes as the starting nodes (i.e., starting variable nodes) are all multi-branch nodes. 
So, we can get the representation of the starting variable node through its branch path sequences. 
Based on the characteristics of the multi-branch structure, we call the process of obtaining the representation of the multi-branch variable node from its branch path sequences as the \textbf{fork query}. 
As shown in Figure \ref{figure1} (B), 
by encoding fork queries, we can get the representation of the multi-branch variable node to complete subsequent path queries, based on which we can independently and recursively encode all subsequent path query sequences to get the representation of the free variable. 

In this way, a tree-like EFOL query can be decomposed into path queries and fork queries so that only two encoders, a path query encoder and a fork query encoder, 
are needed to encode the entire query computation tree to obtain the query embedding, namely the representation of the free variable.

\subsection{Path Query Encoder}
\label{Path Query Encoder}

Clearly, a path query can be regarded as a sequence. 
It contains a starting node $n_{s}$, one or more unknown variable nodes 
$n_{v_{1}},n_{v_{2}},...,n_{v_{k}}$ and a corresponding sequence of set operators $\{l_{1},l_{2},...,l_{k}\}$ where 
$k\ge 1$ and $l_{i}$ is a relational projection operator or negation operator. 
A path query can be expressed as: 
\begin{equation}
    n_{s}\stackrel{l_{1}}{\longrightarrow}n_{v_{1}}\stackrel{l_{2}}{\longrightarrow}n_{v_{2}}\ \cdots \ n_{v_{k-1}}\stackrel{l_{k}}{\longrightarrow}n_{v_{k}} \\\label{neq1}.
\end{equation}
To model the complex dependencies between various parts of a path query, 
we use the encoder part of the transformer \cite{vaswani2017attention} to implement our path query encoder. 
Through the bidirectional attention mechanism, we can make full use of future information during the path query encoding process. 
However, the path query contains unknown variable nodes
$n_{v_{i}}$, so it cannot be encoded directly. 
To obtain a sequence that can be used as input to the path query encoder, we represent the path query with its set operator sequence and known starting node. 
Specifically, a starting node followed by a sequence of set operators uniquely determines a path query, namely: 
\begin{equation}
    n_{s},l_{1},l_{2},...,l_{k}\Rightarrow n_{s}\stackrel{l_{1}}{\longrightarrow}n_{v_{1}}\stackrel{l_{2}}{\longrightarrow}n_{v_{2}}\ \cdots \ n_{v_{k-1}}\stackrel{l_{k}}{\longrightarrow}n_{v_{k}} \\\label{neq2}.
\end{equation}
For the starting node \(n_{s}\), if it is an anchor node, we can represent it with the corresponding entity embedding, and if it is a multi-branch variable node, we can obtain its representation through the fork query. 
For the set operator sequence 
$\{l_{1},l_{2},...,l_{k}\}$, it involves two kinds of set operators: relational projection and negation. Relational projection operators correspond to relations one-to-one, so we can directly use relation embeddings to represent the corresponding relational projection operators. 
However, the negation operator has no specific corresponding representation. 
Therefore, in order to encode the entire path query with negation at one time, we introduce a token $\mathrm{[Negation]}$ to represent the set negation operation. For example, $\mathrm {Path\ Query\ 2}$ in Figure \ref{figure2} can be expressed as the following sequence: $\{ e_{2},r_{2},\mathrm{[Negation]}\} $. 
After embedding the $\mathrm{[Negation]}$ token, 
the anchor entities, and the corresponding relations for each relational projection operator into the $d$-dimensional vector space, 
all elements in the sequence will have corresponding representations, 
so we can obtain the corresponding embedding sequences as the encoder input. 
As shown in Figure \ref{figure2}, 
given a path query, we first transform it into the corresponding input embedding sequence $E_{in}$: $\{E_{1},E_{2},...,E_{m}\}$ where each $E_{i}\in \mathbb{R}^{d}$, 
then we feed it into the encoder defined by Equation \ref{eq1} to get the corresponding path query embedding $E_{pq}\in \mathbb{R}^{d}$: 
\begin{equation}
    E_{pq}=\mathrm {MP}(\mathrm {TrmE_{k_{1}}}(E_{in})) \\\label{eq1},
\end{equation}
where $\mathrm {TrmE_{k_{1}}}$ is a transformer encoder with $k_{1}$ layers, and $\mathrm {MP}$ is a mean pooling layer.

\subsection{Fork Query Encoder}
\label{Fork Query Encoder}

For a query computation tree, the fork query encoder is used to aggregate each branch path query embedding of a multi-branch variable node to obtain the embedding of the node. 
For a fork query, there are two set operators: intersection and union. 
Following the previous work \cite{ren2020query2box,ren2020beta,amayuelas2022neural}, 
we also use DNF to handle disjunctive queries, 
so that we only need to consider the intersection operator. 
A branch node may have multiple branches, such as a three-branch node, 
that is, the intersection operation may have multiple inputs. 
Inspired by \cite{amayuelas2022neural}, 
we consider the intersection operation between two inputs here, 
since multiple inputs can be transformed into a pairwise intersection. 
We use a multi-layer perceptron with $k_{2}$ layers as the fork query encoder: 
\begin{equation}
    E_{v}=\mathrm {MLP_{k_{2}}}(E_{pq_{i}}\otimes E_{pq_{j}}) \\\label{eq2},
\end{equation}
where the input $E_{pq_{i}},E_{pq_{j}}\in \mathbb{R}^{d}$ are the branch path query embeddings of the corresponding multi-branch variable node $v$, 
$\otimes$ is the concat operation, 
and the output $E_{v}\in \mathbb{R}^{d}$ is the embedding of the node. 
As shown in Figure \ref{figure2}, after obtaining the embedding of the multi-branch intermediate variable node $V$ by the fork query encoder, we can recursively encode the subsequent path queries.

\subsection{Learning Procedure}
\label{Learning Procedure}

\subsubsection{Distance Function}
To match the correct answer entity and the query, we need to define a distance function. 
In our work, entities and queries are encoded into a $d$-dimensional one-point vector space \cite{hamilton2018embedding, amayuelas2022neural}, 
rather than special vector spaces such as box embedding space or probability distribution embedding space as in some previous work \cite{ren2020query2box, ren2020beta}. 
Therefore, the simple Euclidean distance is good enough for us to calculate the distance between two points in an Euclidean space. 
Given a query $q$ and a candidate entity $e\in \mathcal{V}$, 
after getting the corresponding embeddings 
$v_{q},v_{e}\in \mathbb{R}^{d}$, 
we have the following functions, where $|.|$ is $L1$ norm: 
\begin{equation}
    \mathrm {Dist}(e,q)=|v_{e}-v_{q}| \\\label{eq3}.
\end{equation}

\subsubsection{Training Objective}
The goal is to jointly train the query encoders and the embeddings of entities, relations, and the negation token. 
Our training objective is to minimize the distance between the query embedding and the corresponding answer entity embeddings, 
while maximizing the distance between the query embedding and the incorrect random entity embeddings obtained from the negative sampling strategy \cite{sun2019rotate}, 
which we define as follows: 
\begin{equation}
    \mathcal{L} =-\mathrm {log}\sigma (\gamma - \mathrm {Dist}(e;q))-\sum_{j=1}^{u}\frac{1}{u}  \mathrm {log}\sigma (\mathrm {Dist}(e_{j}';q)-\gamma ) \\\label{eq4},
\end{equation}
where $\sigma$ is the sigmoid function, 
$q$ is the query, $e \in \llbracket q \rrbracket$ is the answer entity of the corresponding query (i.e., the positive sample), 
$e'_{j}\notin \llbracket q \rrbracket$ is the $j$-th random negative sample (non-answer to the query $q$), 
$\gamma$ represents a fixed margin, 
and $u$ is the number of negative samples. 
Both $\gamma$ and $u$ are hyperparameters.

\subsubsection{Inference}
Given a query $q$, 
we use the path query encoder and fork query encoder to encode it by following the query computation tree to obtain the query embedding $v_{q}$. 
Based on the distance defined in Equation \ref{eq3} from $v_{q}$ to all entity embeddings, 
all entities are then ranked via near-neighbor search using Locality Sensitivity Hashing \cite{indyk1998approximate}.  

\begin{table*}[htbp]
  \centering
  \caption{The MRR results (\%) of baselines and Pathformer on answering EPFO queries. The \textbf{boldface} indicates the best results, and the second-best ones are marked with \underline{underlines}.}
    \resizebox{0.7\linewidth}{!}{
    \begin{tabular}{cccccccccccc}
    \toprule
    \textbf{Dataset} & \textbf{Model} & \textbf{1p} & \textbf{2p} & \textbf{3p} & \textbf{2i} & \textbf{3i} & \textbf{ip} & \textbf{pi} & \textbf{2u} & \textbf{up} & \textbf{avg} \\
    \midrule
    \multirow{10}[2]{*}{\textbf{FB15k-237}} & \textbf{GQE} & 35.0    & 7.2   & 5.3   & 23.3  & 34.6  & 10.7  & 16.5  & 8.2   & 5.7   & 16.3 \\
          & \textbf{Q2B} & 40.6  & 9.4   & 6.8   & 29.5  & 42.3  & 12.6  & 21.2  & 11.3  & 7.6   & 20.1 \\
          & \textbf{BetaE} & 39.0    & 10.9  & 10.0    & 28.8  & 42.5  & 12.6  & 22.4  & 12.4  & 9.7   & 20.9 \\
          & \textbf{Q2P} & 39.1  & 11.4  & 10.1  & 32.3  & \underline{47.7}  & 14.3  & 24.0    & 8.7   & 9.1   & 21.9 \\
          & \textbf{ConE} & 41.5  & 11.7  & \textbf{10.6}  & 30.9  & 45.4  & 12.1  & 23.6  & 13.7  & 9.9   & 22.2 \\
          & \textbf{MLP} & 42.7  & \underline{12.4}  & \textbf{10.6}  & 31.7  & 43.9  & 14.9  & 24.2  & 13.7  & 9.7   & 22.6 \\
          & \textbf{MLP-Mixer} & 42.4  & 11.5  & 9.9   & 33.5  & 46.8  & 14.0    & 25.4  & 14.0    & 9.2   & 22.9 \\
          & \textbf{FuzzQE} & \underline{42.8}  & \textbf{12.9}  & \underline{10.3}  & 33.3  & 46.9  & \textbf{17.8}  & \textbf{26.9}  & \underline{14.6}  & \textbf{10.3}  & \underline{24.0} \\
          & \textbf{GammaE} & 40.1  & 11.6  & \underline{10.3}   & \textbf{35.6}  & \textbf{49.5}  & 12.4  & 25.1  & \textbf{14.9}  & \underline{10.2}   & 23.3 \\
          & \textbf{Pathformer} & \textbf{44.8}  & \textbf{12.9}  & \textbf{10.6}  & \underline{34.2}  & 47.3  & \underline{17.0}    & \underline{26.2}  & \textbf{14.9}  & 10.0    & \textbf{24.2} \\
    \midrule
    \multirow{10}[2]{*}{\textbf{NELL995}} & \textbf{GQE} & 32.8  & 11.9  & 9.6   & 27.5  & 35.2  & 14.4  & 18.4  & 8.5   & 8.8   & 18.6 \\
          & \textbf{Q2B} & 42.2  & 14.0    & 11.2  & 33.3  & 44.5  & 16.8  & 22.4  & 11.3  & 10.3  & 22.9 \\
          & \textbf{BetaE} & 53.0    & 13.0    & 11.4  & 37.6  & 47.5  & 14.3  & 24.1  & 12.2  & 8.5   & 24.6 \\
          & \textbf{Q2P} & \textbf{56.5}  & 15.2  & 12.5  & 35.8  & 48.7  & \textbf{22.6}  & 16.1  & 11.1  & 10.4  & 25.5 \\
          & \textbf{ConE} & 53.0    & 15.2  & 13.2  & \underline{39.6}  & \underline{50.5}  & 16.8  & 25.4  & \underline{14.8}  & 10.9  & 26.6 \\
          & \textbf{MLP} & 55.2  & 16.8  & \textbf{14.9}  & 36.4  & 48.0    & 18.2  & 22.7  & 14.7  & \underline{11.3}  & 26.5 \\
          & \textbf{MLP-Mixer} & 55.4  & 16.5  & 13.9  & 39.5  & \textbf{51.0}    & 18.3  & 25.7  & 14.7  & 11.2  & \underline{27.4} \\
          & \textbf{FuzzQE} & 47.4  & \underline{17.2}  & \underline{14.6}  & 39.5  & 49.2  & \underline{20.6}  & \textbf{26.2}  & \textbf{15.3}  & \textbf{12.6}  & 27.0 \\
          & \textbf{GammaE} & 53.0      & 12.8      & 11.4      & 37.1      & 48.6      & 13.4      & 23.6      & 11.5      & 8.5      & 24.4 \\
          & \textbf{Pathformer} & \underline{56.4}  & \textbf{17.4}  & \textbf{14.9}  & \textbf{39.9}  & 50.4  & 19.4  & \underline{26.0}    & 14.4  & 11.1  & \textbf{27.8} \\
    \bottomrule
    \end{tabular}%
    }
  \label{table1}%
\end{table*}%

\section{Experiments}

\subsection{Datasets and Experiment Setup}

\subsubsection{Datasets and Queries}
\label{5.1.1}

We consider the CLQA benchmark proposed in BetaE\cite{ren2020beta}, which focuses on incomplete KGs and is widely used in previous works \cite{zhang2021cone, chen2022fuzzy, yang2022gammae, amayuelas2022neural}. 
In this benchmark, the performance measurement is exclusively focused on answer entities that require (implicitly) imputing at least one edge. 
Such answer entities are called non-trivial answers. 
In other words, the goal is to obtain non-trivial answers to arbitrary complex queries that cannot be discovered by directly traversing the KG. 
For this purpose, 
the edges (i.e., triples) in each dataset of the benchmark are divided into training, validation, and test edges, from which three graphs can be obtained: $\mathcal{G} _{\text { train}}$, $\mathcal{G} _{\text {valid}}$, and $\mathcal{G} _{\text {test}}$. 
Specifically, $\mathcal{G} _{\text { train}}$ only consists of training edges, $\mathcal{G} _{\text {valid}}$ consists of training and validation edges, and 
$\mathcal{G} _{\text {test}}$ contains all edges, namely   $\mathcal{G} _{\text {train}}\subseteq \mathcal{G} _{\text {valid}}\subseteq \mathcal{G} _{\text {test}}$. 
Given a query $q$, $\llbracket q \rrbracket_{\text {train }}$, $\llbracket q \rrbracket_{\text {valid }}$, and $\llbracket q \rrbracket_{\text {test }}$ represent the answer entities sets  obtained on $\mathcal{G} _{\text {train}}$, $\mathcal{G} _{\text {valid}}$, and $\mathcal{G} _{\text {test}}$, respectively. 
Obviously, these sets follow $\llbracket q \rrbracket_{\text {train }} \subseteq \llbracket q \rrbracket_{\text {valid }}\subseteq \llbracket q \rrbracket_{\text {test }}$. 
Since our evaluation concentrates on incomplete KGs,
for a given test (validation) query $q$, our primary interest lies in obtaining non-trivial answers $ \llbracket q \rrbracket_{\text {test }} \backslash \llbracket q \rrbracket_{\text {valid }}$ ($\llbracket q \rrbracket_{\text {valid }} \backslash \llbracket q \rrbracket_{\text {train }}$). 
We conduct experiments on two KG datasets of the benchmark: FB15k-237 \cite{toutanova2015observed} and NELL995 \cite{xiong2017deeppath}. 

\begin{figure}
    \centering
    \includegraphics[width=0.9\linewidth]{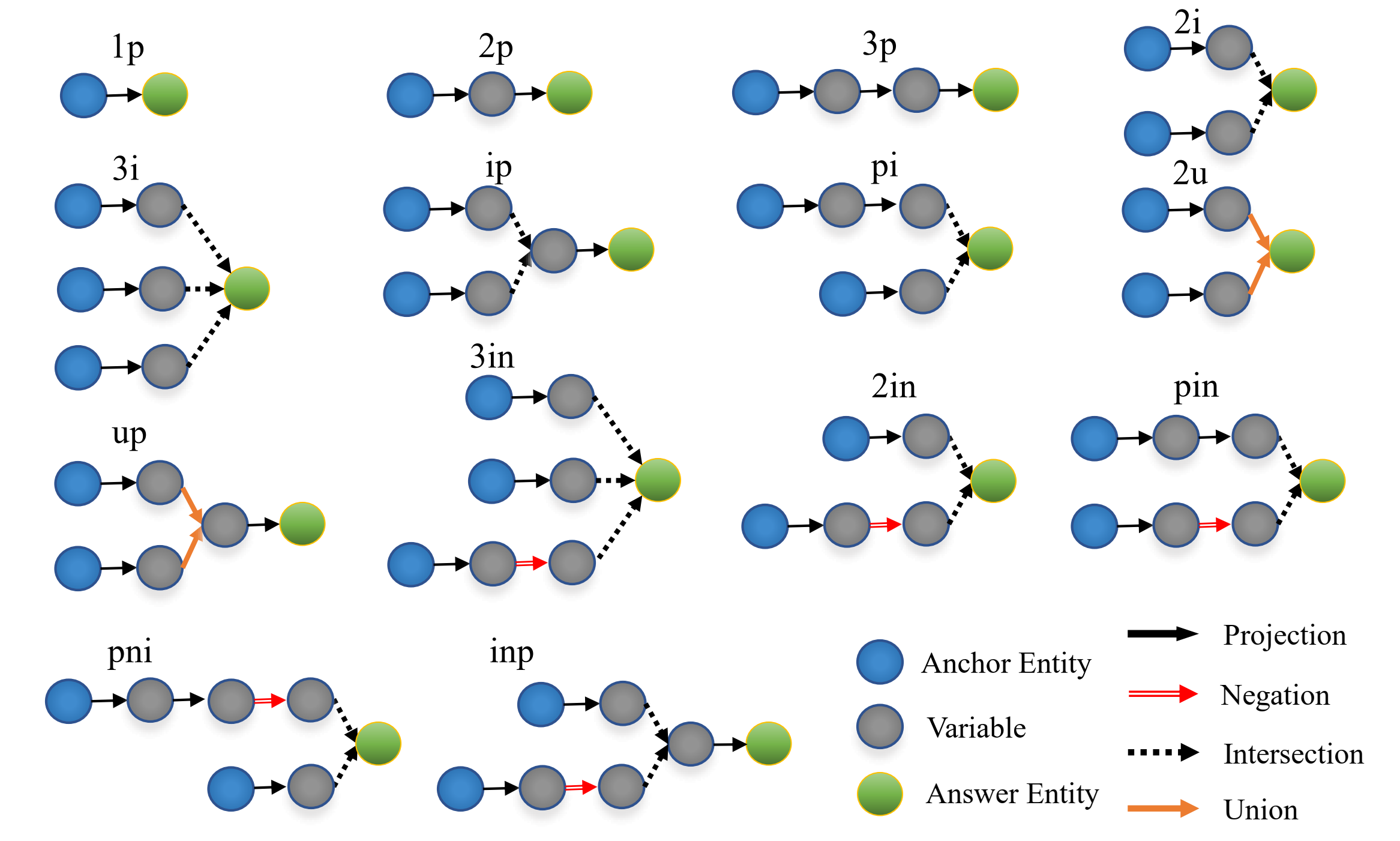}
    \caption{All the query structures used in our experiments, where $p$, $i$, $u$, and $n$ represent relational projection, intersection, union, and negation, respectively. }
    \label{figure3}
\end{figure}

Specifically, the benchmark contains 14 query structures, as shown in Figure \ref{figure3}.  
The training and evaluation queries contain five conjunctive query structures ($1p/2p/3p/2i/3i$) and five query structures with negation ($2in/3in/inp/pni/pin$). 
To evaluate the further generalization ability of the model, the other four query structures ($ip/pi/2u/up$) are used only for evaluation but not for training. That is, they are zero-shot queries. 
It is worth noting that the existing methods evaluate their performance on two sets of complex logical queries: 
existential positive first order (EPFO) queries and existential first order logic (EFOL) queries, 
where EPFO queries do not involve the negation operator. 
They typically evaluate the model's ability to answer queries without negation on EPFO queries, where the model is trained on only five conjunctive query structures. 
When evaluating the model's ability on queries with negation, the model is trained on the full EFOL queries, namely using all 10 query structures for training and evaluation. 
To have a comparable result, 
our experiments adopt the same strategy. 
In general, we conduct experiments on the BetaE datasets.  
However, the BIQE \cite{kotnis2021answering} in the baselines (it will be later discussed in Section \ref{5.1.3}) we consider for comparison is trained and evaluated on conjunctive queries on the Query2box (Q2B) datasets \cite{ren2020query2box}. 
To make a fair comparison with BIQE, we also conduct experiments on the FB15k-237 and NELL995 configured by the Q2B setting. We denote these two datasets as FB237-Q2B and NELL-Q2B. 
One difference between the BetaE datasets and the Q2B datasets is the limit on the number of answers to the queries, since the Q2B datasets have queries with more than 5000 answers, almost $1/3$ of some datasets. Furthermore, the Q2B datasets do not contain queries with negation. 
Thus, we compare Pathformer and BIQE on the Q2B datasets. 
As for all the other experiments, we perform on the BetaE datasets. 
More details about the datasets can be found in Supplementary Material \ref{app.A}. 

\subsubsection{Evaluation Protocol}
\label{eva}

For each non-trivial answer $e \in \llbracket q \rrbracket_{\text {test }} \backslash \llbracket q \rrbracket_{\text {valid }}$ corresponding to a test query $q$, 
we rank $e$ against non-answer entities $\mathcal{V} \backslash \llbracket q \rrbracket_{\text {test}}$ based on the distance defined in Equation \ref{eq3}. 
Given the rank $rank(e)$ of the answer entity $e$, 
we can calculate the evaluation metric for answering the query $q$: 
\begin{equation}
    \mathrm {Metric} (q)=\frac{1}{|\llbracket q \rrbracket_{\text {test }} \backslash \llbracket q \rrbracket_{\text {valid }}|} \sum_{e \in \llbracket q \rrbracket_{\text {test }} \backslash \llbracket q \rrbracket_{\text {valid }}}^{} f_{\text {metrics}}(\mathrm {rank} (e)). \\\label{eq5}
\end{equation}
In this paper, we follow previous works \cite{amayuelas2022neural,ren2020beta,yang2022gammae} and report the Mean Reciprocal Rank (MRR) results, 
so $f_{metrics}(x)=\frac{1}{x}$. 
We take the MRR average of all queries in the same query structure separately and report the results separately for each query structure. 
In the validation phase, we used the same evaluation metric to evaluate on $\llbracket q \rrbracket_{\text {valid }} \backslash \llbracket q \rrbracket_{\text {train }}$.

\subsubsection{Baselines and Model Variants}
\label{5.1.3}

We consider several competitive neural methods for CLQA tasks as our baselines to compare: 
GQE \cite{hamilton2018embedding}, Q2B \cite{ren2020query2box}, BetaE \cite{ren2020beta}, 
BIQE \cite{kotnis2021answering}, 
Query2particles (Q2P) \cite{bai2022query2particles}, 
ConE \cite{zhang2021cone}, MLP \cite{amayuelas2022neural}, MLP-Mixer \cite{amayuelas2022neural}, FuzzQE \cite{chen2022fuzzy}, and GammaE \cite{yang2022gammae}, 
where GQE, Q2B, BIQE, and MLP-Mixer can only handle EPFO queries, while other methods can handle queries with negation. 
Inspired by \cite{amayuelas2022neural}, 
we also modify the original Pathformer to improve the initial results potentially. 
We consider the following modifications: 
\begin{itemize}
\item \emph{Mixer:} The fork query encoder in the original Pathformer adopts MLP. Here, we consider replacing it with MLP-Mixer \cite{tolstikhin2021mlp}, which may boost the performance of Pathformer on queries involving intersection operators since MLP-Mixer performs better than MLP on average on these query structures according to \cite{amayuelas2022neural}. 

\item \emph{MLP2Vector:} Following the MLP method \cite{amayuelas2022neural}, in order to increase the robustness of the encoder, we use two separate MLPs to calculate the fork query embedding separately and then average their outputs as the final fork query embedding. 
\end{itemize}
For both the main experiments and the model variant experiments, 
our path query encoder uses a six-layer transformer encoder. 
For a fair comparison, we follow previous works \cite{amayuelas2022neural,chen2022fuzzy,ren2020beta} and set our embedding dimension \(d = 800\), and retrain some baselines with different embedding dimensions so that the embedding dimension $d$ of all methods is $800$. 
For GammaE, its union network has parameters that need to be learned, and the query structures involving the union operator in the experiments are all zero-shot queries. 
Therefore, to be fair, we use DNF treatment to compare union operators. 
In addition, BIQE only evaluates conjunctive queries and does not evaluate $2u$ and $up$, so we reimplement it on FB237-Q2B and NELL-Q2B for comparison. 
For more experimental details, please refer to Supplementary Material \ref{app.B}. 

\begin{table*}[htbp]
  \centering
  \caption{The MRR results (\%) of BIQE and Pathformer in answering EPFO queries on the Q2B datasets. The \textbf{boldface} indicates the best results. }
    \resizebox{0.7\linewidth}{!}{
    \begin{tabular}{cccccccccccc}
    \toprule
    \textbf{Dataset} & \textbf{Model} & \textbf{1p} & \textbf{2p} & \textbf{3p} & \textbf{2i} & \textbf{3i} & \textbf{ip} & \textbf{pi} & \textbf{2u} & \textbf{up} & \textbf{avg} \\
    \midrule
    \multirow{2}[2]{*}{\textbf{FB237-Q2B}} & \textbf{BIQE} & 41.3  & 26.3  & 22.8  & 29.5  & 41.5  & 10.7  & 16.1  & 19.2  & 17.9  & 25.0 \\
          & \textbf{Pathformer} & \textbf{44.7}  & \textbf{28.0}    & \textbf{23.7}  & \textbf{33.0}    & \textbf{44.4}  & \textbf{14.3}  & \textbf{22.5}  & \textbf{27.7}  & \textbf{18.5}  & \textbf{28.5} \\
    \midrule
    \multirow{2}[2]{*}{\textbf{NELL-Q2B}} & \textbf{BIQE} & \textbf{58.0}    & 28.1  & \textbf{31.4}  & 31.9  & 46.4  & 8.1   & 15.7  & 31.5  & 18.4  & 29.9 \\
          & \textbf{Pathformer} & 56.1  & \textbf{31.9}  & 30.7  & \textbf{35.9}  & \textbf{49.4}  & \textbf{16.1}  & \textbf{23.8}  & \textbf{41.8}  & \textbf{21.8}  & \textbf{34.2} \\
    \bottomrule
    \end{tabular}%
    }
  \label{table2}%
\end{table*}%

\subsection{Main Results}

Table \ref{table1} shows the MRR results of Pathformer and other competitive neural QE methods in answering EPFO queries on the BetaE datasets. 
The results show that Pathformer has a better average performance on EPFO queries on the two benchmark datasets, reflecting the effectiveness of Pathformer. 
It is worth noting that Pathformer also achieves good results on $1p$ queries. 
According to the discussion in \cite{guu2015traversing}, 
at a high level, it can be considered that training on complex logical queries can provide some beneficial structural regularization. 
And the paths in a KG have proven to be important features for predicting the existence of single edges \cite{neelakantan2015compositional}, 
so training on complex logical queries can enhance the performance of the model on one-hop queries (i.e., $1p$ queries). 
Therefore, we speculate that Pathformer's better performance on complex queries with other structures also potentially enhances Pathformer's performance on $1p$ queries. 

Table \ref{table2} shows the MRR results of BIQE and Pathformer in answering EPFO queries on the Q2B datasets. 
The results show that Pathformer outperforms BIQE on the two benchmark datasets, 
which can verify our speculation to a certain extent. That is, 
only introducing positional encoding to model the graph structure is insufficient, 
which may result in the loss of the input query graph structure information, thereby negatively impacting the performance of BIQE. 
On the other hand, it is not clear how to introduce appropriate graph inductive biases for the computation graph with logical dependencies. 
In contrast, Pathformer only uses the transformer to model the path query sequences, thereby encoding the query computation tree path by path,
avoiding the loss of input information caused by insufficient graph inductive biases for the transformer. 
Moreover, Pathformer can also handle negative queries naturally, whereas BIQE only considers EPFO queries. 

For negative queries, Table \ref{table3} shows the MRR results of Pathformer and other baselines that support negative operator on the BetaE datasets in answering negative queries. 
Judging from the experimental results, Pathformer can support negation by simply introducing a token to represent the negation operator. Although Pathformer does not achieve optimal results on negative queries, it is still better than most competitive baselines overall. 
Combined with Pathformer's excellent performance on EPFO queries, our conclusion is still valid. 

\begin{table}[htbp]
  \centering
  \caption{The MRR results (\%) of baselines and Pathformer on answering queries with negation. The \textbf{boldface} indicates the best results, and the second-best ones are marked with \underline{underlines}.}
  \resizebox{0.95\linewidth}{!}{
    \begin{tabular}{cccccccc}
    \toprule
    \textbf{Dataset} & \textbf{Model} & \textbf{2in} & \textbf{3in} & \textbf{inp} & \textbf{pin} & \textbf{pni} & \textbf{avg} \\
    \midrule
    \multirow{7}[2]{*}{\textbf{FB15k-237}} & \textbf{BetaE} & 5.1   & 7.9   & 7.4   & 3.6   & 3.4   & 5.5 \\
          & \textbf{Q2P} & 4.4   & 9.7   & 7.5   & 4.6   & 3.8   & 6.0 \\
          & \textbf{ConE} & 5.5   & 9.0     & 7.2   & 4.2   & 3.6   & 5.9 \\
          & \textbf{MLP} & \underline{6.6}   & \underline{10.7}  & \underline{8.1}   & \underline{4.7}   & \underline{4.4}   & 6.9 \\
          & \textbf{FuzzQE} & \textbf{8.5}   & \textbf{11.6}  & 7.8   & \textbf{5.2}   & \textbf{5.8}   & \textbf{7.8} \\
          & \textbf{GammaE} & 4.8   & 8.6   & 7.4   & 4.5   & 3.0     & 5.7 \\
          & \textbf{Pathformer} & 6.4   & \textbf{11.6}  & \textbf{8.3}   & \underline{4.7}   & \underline{4.4}   & \underline{7.1} \\
    \midrule
    \multirow{7}[2]{*}{\textbf{NELL995}} & \textbf{BetaE} & 5.1   & 7.8   & 10.0    & 3.1   & 3.5   & 5.9 \\
          & \textbf{Q2P} & 5.1   & 7.4   & 10.2  & 3.3   & 3.4   & 6.0 \\
          & \textbf{ConE} & \underline{5.6}   & 8.1   & \underline{10.8}  & 3.6   & \underline{3.9}   & \underline{6.4} \\
          & \textbf{MLP} & 5.1   & 8.0     & 10.0    & 3.6   & 3.6   & 6.1 \\
          & \textbf{FuzzQE} & \textbf{7.8}   & \textbf{9.8}   & \textbf{11.1}  & \textbf{4.9}   & \textbf{5.5}   & \textbf{7.8} \\
          & \textbf{GammaE} & 4.2   & 6.5   & 9.7   & 3.4   & 3.0     & 5.4 \\
          & \textbf{Pathformer} & 5.1   & \underline{8.6}   & 10.3  & \underline{3.9}   & 3.5   & 6.3 \\
    \bottomrule
    \end{tabular}%
    }
  \label{table3}%
\end{table}%

We evaluate the model variants of Pathformer on the EPFO queries of the BetaE datasets, and the average MRR results are shown in Table \ref{table4}. 
For the Mixer variant, its performance is almost the same as that of the original Pathformer, but the MLP-Mixer \cite{tolstikhin2021mlp} has fewer parameters than MLP, so it may have an advantage over MLP in some situations. 
MLP2vector approach can significantly improve the performance of the MLP method in \cite{amayuelas2022neural}, but from our experimental results, 
MLP2vector does not enhance the performance of Pathformer. 
Just like the analyses in \cite{amayuelas2022neural}, this is because the MLP method does not reach the optimal solution for encoding the branch path queries, so using MLP2vector with more parameters as the intersection network can help provide better solutions. 
However, Pathformer obtains a relatively good enough representation by encoding branch path queries with a standard transformer encoder, so the MLP2vector variant does not yield performance improvements. 
For the MRR results of the variants on each query structure, please refer to Supplementary Material \ref{app.B}.

\begin{table}[htbp]
  \centering
  \caption{The average MRR results (\%) of the model variants of Pathformer on answering EPFO queries. }
  \resizebox{0.8\linewidth}{!}{
    \begin{tabular}{ccc}
    \toprule
    \multirow{2}[4]{*}{\textbf{Model Variant}} & \multicolumn{2}{c}{\textbf{avg}} \\
\cmidrule{2-3}          & \textbf{FB15k-237} & \textbf{NELL995} \\
    \midrule
    \textbf{Pathformer-Mixer} & 24.3      & 27.7 \\
    \textbf{Pathformer-MLP2vector} & 24.3      & 27.7 \\
    \bottomrule
    \end{tabular}%
    }
  \label{table4}%
\end{table}%

\subsection{Does Bidirectional Attention Help Model Implicit Dependencies?}

To verify whether the future context information generated by the bidirectional attention mechanism can help Pathformer model the implicit dependencies between different parts of a path query, we use masked attention to implement the path query encoder. 
In other words, the future context information is not considered in the path query encoding process in this case. 
We use this variant to compare with the original Pathformer and use a one-layer transformer encoder to conduct this experiment on the EPFO queries of the FB15k-237 dataset. 
The results are shown in Table \ref{table5}. The performance of Pathformer without future context information has a significant drop. 
As described above, pathformer shares a similar spirit with BIQE in the sense that both methods use bidirectional attention to acquire future information to model the implicit complex dependencies between various parts of a query. 
BIQE also conducted experiments to verify this motivation. Based on BIQE's experimental analyses \cite{kotnis2021answering} and our experimental results, we believe that the query context information introduced by bidirectional attention is indeed helpful in modeling implicit dependencies.

\begin{table}[htbp]
  \centering
  \caption{The average MRR results (\%) on answering EPFO queries. The number in parentheses represents the number of layers of the path query encoder. }
  \resizebox{0.6\linewidth}{!}{
    \begin{tabular}{cc}
    \toprule
    \textbf{Model} & \textbf{avg} \\
    \midrule
    \textbf{Pathformer(No future context)(1)} & 22.4 \\
    \textbf{Pathformer(1)} & 23.2 \\
    \bottomrule
    \end{tabular}%
    }
  \label{table5}%
\end{table}%

\subsection{More analyses on Pathformer}

For the path query sequence, its elements are strictly ordered. 
To encode the order information of the input query path, we consider the absolute position encoding of the original transformer. 
The transformer's position encoding is a widely proven experimentally valid way of injecting order prior information. 
In addition, we also consider the relative position encoding \cite{shaw2018self}. 
We conduct experiments on position encoding on the EPFO queries of FB15k-237, and the results are shown in Table \ref{PE}, where "w/o PE" indicates "without the positional encoding" and "w/ RPE" indicates "with the relative positional encoding". 
Experimental results show that using positional encoding is better than not using them, indicating that these positional encoding schemes are able to learn the order information of path query sequences. 
Although the relative position encoding scheme is slightly better than the original absolute position encoding scheme, 
we use absolute position encoding as our default scheme because the additional computational cost is caused by relative position encoding.

\begin{table}[htbp]
  \centering
  \caption{The average MRR results (\%) on answering EPFO queries of FB15k-237. }
  \resizebox{0.4\linewidth}{!}{
    \begin{tabular}{cc}
    \toprule
    \textbf{Model} & \textbf{avg} \\
    \midrule
    \textbf{Pathformer} & 24.2 \\
    \textbf{Pathformer w/o PE} & 24.1 \\
    \textbf{Pathformer w/ RPE} & 24.4 \\
    \bottomrule
    \end{tabular}%
    }
  \label{PE}%
\end{table}%

Since Pathformer introduces the transformer, 
it may require more parameters than some other neural QE methods. 
However, when Pathformer reduces the number of layers of the path query encoder to shorten the distance between the model parameters and those of most baselines, 
it can still outperform these baselines. 
As shown in Table \ref{layers}, when we use only a one-layer transformer encoder, 
the average MRR result of Pathformer on EPFO queries is $23.2$. 
The result can still perform better than most baselines on FB15k-237. 
In particular, 
as a neural one-point embedding method, the most relevant baseline to our work is the MLP method \cite{amayuelas2022neural}, which is also a one-point embedding method. 
Pathformer differs from it in that we introduce bidirectional attention \cite{vaswani2017attention} to process all logical operators on one path at a time, while MLP uses different neural networks to model projection operation and negation operation link by link according to the computation graph. 
Therefore, here we focus on comparing with the MLP method. 
As shown in Table \ref{layers}, when we use only a one-layer transformer encoder, 
the parameters of our model are close to those of the MLP method. 
However, our method still performs significantly better by comparison, 
suggesting that using bidirectional self-attention to process the entire path at once can better model path queries. 
Furthermore, we can observe that the model achieves better performance as the number of transformer encoder layers increases, 
as the number of parameters of the model also increases, 
allowing us to model better the implicit complex dependencies between parts of a path query. For the MRR results on each query structure for Pathformer with the different number of transformer encoder layers, please refer to Supplementary Material \ref{app.B}. 
For more analyses on the time computational costs, please refer to Supplementary Material \ref{costsss}. 

\begin{table}[htbp]
  \centering
  \caption{The average MRR results (\%) of EPFO queries on FB15k-237 for Pathformer with the different number of transformer encoder layers. The number in parentheses represents the number of layers of the transformer encoder. The parameters in the table are in millions. }
  \resizebox{0.55\linewidth}{!}{
    \begin{tabular}{ccc}
    \toprule
    \textbf{Model} & \textbf{avg} & \textbf{Parameters} \\
    \midrule
    \textbf{MLP} & 22.6  & 9.0 \\
    \textbf{Pathformer(1)} & 23.2  & 12.2 \\
    \textbf{Pathformer(2)} & 23.7  & 19.8 \\
    \textbf{Pathformer(4)} & 24.1  & 35.2 \\
    \textbf{Pathformer(6)} & 24.2  & 50.6 \\
    \textbf{Pathformer(8)} & 24.4  & 66.0 \\
    \bottomrule
    \end{tabular}%
    }
  \label{layers}%
\end{table}%

\subsection{Pathformer Combined with Some Other Neural Methods}
\label{combine}

Here, we combine the Pathformer with some other neural QE methods. 
Specifically, Pathformer is a neural one-point embedding method because it embeds entities and queries into a one-point vector space. 
To combine with some other neural QE methods, we generalize Pathformer to other embedding spaces, such as box embedding space \cite{ren2020query2box} and beta embedding space \cite{ren2020beta}. 
By embedding entities and queries into a specific vector space with corresponding regularizations and replacing the fork query encoder with the corresponding intersection network, we can combine Pathformer with some other neural QE methods. 
Here, we consider two neural QE methods: Q2B \cite{ren2020query2box} and BetaE \cite{ren2020beta}. 
To verify whether Pathformer can combine these QE methods and enhance their performance, we only use the one-layer transformer encoder as our path query encoder to make the combined model size as close as possible to the corresponding model. 
We train and evaluate these models on the EPFO queries of BetaE datasets, and the experimental results are shown in Table \ref{pathformerBoxBeta}. 
From the experimental results, we can observe that Pathformer can combine with these neural QE methods and enhance their original performance to some extent. These results show that Pathformer has the potential to be applied to non-one-point embedding spaces. 
For more details about this part of the experiment, please refer to Supplementary Material \ref{app.B}.

\begin{table}[htbp]
  \centering
  \caption{The average MRR results (\%) on answering EPFO queries. The number in parentheses represents the number of layers of the path query encoder. }
  \resizebox{0.9\linewidth}{!}{
    \begin{tabular}{ccc}
    \toprule
    \multirow{2}[4]{*}{\textbf{Model}} & \multicolumn{2}{c}{\textbf{avg}} \\
\cmidrule{2-3}          & \textbf{FB15k-237} & \textbf{NELL995} \\
    \midrule
    \textbf{Q2B} & 20.1  & 22.9 \\
    \textbf{Pathformer(1)+Box embedding} & 20.6  & 25.0 \\
    \midrule
    \textbf{BetaE} & 20.9  & 24.6 \\
    \textbf{Pathformer(1)+Beta embedding} & 21.5  & 24.4 \\
    \bottomrule
    \end{tabular}%
    }
  \label{pathformerBoxBeta}%
\end{table}%

\section{Conclusion}

In this paper, we propose Pathformer, a neural QE method for CLQA tasks. 
It decomposes a tree-like EFOL query into path queries and fork queries and then uses the transformer encoder to encode the path query to model the implicit complex dependencies between various parts of the path query. Pathformer introduces the transformer naturally, 
without the need to consider the graph structure modelling problems caused by insufficient graph inductive biases for the transformer. 
Furthermore, by introducing just one token representing negation, Pathformer can encode path queries with negation. 
By using the neural network to encode fork queries, Pathformer can recursively encode all path queries to get the final query embedding. 
Experimental results show that Pathformer outperforms existing competitive neural QE methods, demonstrating its effectiveness. 
Furthermore, after experimental analyses, we find that Pathformer, as a one-point embedding method, has the potential to be applied to non-one-point embedding space. 

\section{Limitation}

Based on query computation tree decomposition, Pathformer decomposes the tree-like query into path query sequences by branches to encode these sequences recursively. Therefore, Pathformer only supports tree-like queries. Although tree-like queries are more in line with the human multi-hop query style, there may still be special cyclic queries that Pathformer cannot handle. 

\balance

\section*{Acknowledgment}
We thank the anonymous reviewers for their helpful feedbacks. 
The work described in this paper was partially funded by the National Natural Science Foundation of China (Grant Nos. 62272173, 62273109), the Natural Science Foundation of Guangdong Province (Grant Nos. 2024A1515010089, 2022A1515010179), the Science and Technology Planning Project of Guangdong Province (Grant No. 2023A0505050106), and the National Key R\&D Program of China (Grant No. 2023YFA1011601).

\bibliography{references}
\bibliographystyle{IEEEtran}

\clearpage

\section*{Supplementary Material}

\subsection{Dataset Statistics}
\label{app.A}

The statistics of the KGs we used are shown in Table \ref{datasetstatistics}. 
The average number of queries and answers for BetaE datasets is shown in Table \ref{queries} and Table \ref{answers}. 
We briefly introduce the difference between the Q2B datasets and the BetaE datasets in Section \ref{5.1.1}. For more details about the Q2B datasets, please refer to the Q2B paper\cite{ren2020query2box}.

\subsection{Experimental Details}
\label{app.B}

Our code is implemented using PyTorch. 
For both the main experiments and the model variant experiments, we adopt the following parameter settings for all our models: 
Embedding dim d = 800, learning rate = 0.0001, negative sample size = 128, batch size = 512, margin = 24, number of iterations = 300000/450000, the number of transformer encoder layers \(\text{k}_{1}\) = 6, dropout rate = 0.1, transformer's feed-forward network embedding dim \(\text {d}_{\text{ffn}}\) = 4d. 
The MRR results of the variants on each query structure are shown in Table \ref{variants details}. 
The MRR results on each query structure for Pathformer with the different number of transformer encoder layers are shown in Table \ref{layers_detail}. 

While for Pathformer combined Q2B \cite{ren2020query2box} and BetaE \cite{ren2020beta}, as explained in Section \ref{combine}, in order to make as fair a comparison as possible, we set $d$ = 400, margin = 30/60, and the number of transformer encoder layers \(\text{k}_{1}\) = 1. 
Since box embedding has two parameters: offset and center, 
and BetaE also has two parameters per distribution: $\alpha$ and $\beta$, 
we use two path query encoders with input dimension $d$ to model the path query encoding process of the two parameters respectively. Specifically, given two input embedding sequences representing two parameters respectively, namely $E_{in}^{1}$ and $E_{in}^{2}$ where the element dimension in the sequences is $d$, we use the Equation \ref{eq1} to obtain the path query embeddings: 
\begin{equation}
    E_{pq}^{1}=\mathrm {MP}(\mathrm {TrmE_{k_{1}}^{1}}(E_{in}^{1})), \\
\end{equation}
\begin{equation}
    E_{pq}^{2}=\mathrm {MP}(\mathrm {TrmE_{k_{1}}^{1}}(E_{in}^{2})), \\
\end{equation}
where $E_{pq}^{1}$ and $E_{pq}^{2}$ respectively correspond to the path query embedding of two parameters. 
We then modify the fork query encoder part of the original Pathformer and regularize the output of the path query encoders appropriately. Specifically, for the path query embeddings, we make the corresponding regularization processing for them: 
\begin{equation}
    {E_{pq}^{1}}^{'}=\mathrm {Regularizer} (E_{pq}^{1}), \\
\end{equation}
\begin{equation}
    {E_{pq}^{2}}^{'}=\mathrm {Regularizer} (E_{pq}^{2}), \\
\end{equation}
where \(\mathrm {Regularizer}\) represents the regularization corresponding to each method, preventing the outputs of the path query encoders from not conforming to the constraints of the method. Accordingly, for the original fork query encoder defined in Equation \ref{eq2} of Section \ref{Fork Query Encoder}, we modify it as follows: 
\begin{equation}
    E_{v}^{1},E_{v}^{2}=\mathrm {Intersection} ({E_{pq_{1}}^{1}}^{'},{E_{pq_{1}}^{2}}^{'},...,{E_{pq_{i}}^{1}}^{'},{E_{pq_{i}}^{2}}^{'}), \\
\end{equation}
where \(i\ge 2\) denotes the number of sets that need to be intersected and \(\mathrm {Intersection}\) represents the intersection module corresponding to each method. 
By inputting the regularized path query embeddings into this module, we can obtain the embedding of the two parameters of the multi-branch variable node \(v\), namely $E_{v}^{1},E_{v}^{2}$. 
Regarding the experimental results of Pathformer combined with these
QE methods, more detailed MRR results on each query structure
are shown in Table \ref{pathformerBoxBeta_detail}. 

Every single experiment on Pathformer can be run independently on a single NVIDIA GeForce GTX 1080 Ti GPU.

\begin{table*}[htbp]
  \centering
  \caption{Datasets statistics as well as training, validation and test edge splits. }
  \resizebox{0.75\linewidth}{!}{
    \begin{tabular}{ccccccc}
    \toprule
    \textbf{Dataset} & \textbf{Entities} & \textbf{Relations} & \textbf{Training Edges} & \textbf{Val Edges} & \textbf{Test Edges} & \textbf{Total Edges} \\
    \midrule
    \textbf{FB15k-237} & 14,505 & 237   & 272,115 & 17,526 & 20,438 & 310,079 \\
    \midrule
    \textbf{NELL995} & 63,361 & 200   & 114,213 & 14,324 & 14,267 & 142,804 \\
    \bottomrule
    \end{tabular}%
    }
  \label{datasetstatistics}%
\end{table*}%
\begin{table*}[htbp]
  \centering
  \caption{The number of training, validation, and test queries generated for different query structures of the BetaE datasets. }
  \resizebox{0.6\linewidth}{!}{
    \begin{tabular}{ccccccc}
    \toprule
    \textbf{Queries} & \multicolumn{2}{c}{\textbf{Training}} & \multicolumn{2}{c}{\textbf{Validation}} & \multicolumn{2}{c}{\textbf{Test}} \\
    \midrule
    \textbf{Dataset} & 1p/2p/3p/2i/3i & 2in/3in/inp/pin/pni & 1p    & Others & 1p    & Others \\
    \midrule
    \textbf{FB15k-237} & 149,689 & 14,968 & 20,101 & 5,000 & 22,812 & 5,000 \\
    \textbf{NELL995} & 107,982 & 10,798 & 16,927 & 4,000 & 17,034 & 4,000 \\
    \bottomrule
    \end{tabular}%
    }
  \label{queries}%
\end{table*}%
\begin{table*}[htbp]
  \centering
  \caption{The average number of answers divided by query structure for BetaE datasets. }
  \resizebox{0.8\linewidth}{!}{
    \begin{tabular}{ccccccccccccccc}
    \toprule
    \textbf{Dataset} & \textbf{1p} & \textbf{2p} & \textbf{3p} & \textbf{2i} & \textbf{3i} & \textbf{ip} & \textbf{pi} & \textbf{2u} & \textbf{up} & \textbf{2in} & \textbf{3in} & \textbf{inp} & \textbf{pin} & \textbf{pni} \\
    \midrule
    \textbf{FB15k-237} & 1.7   & 17.3  & 24.3  & 6.9   & 4.5   & 17.7  & 10.4  & 19.6  & 24.3  & 16.3  & 13.4  & 19.5  & 21.7  & 18.2 \\
    \midrule
    \textbf{NELL995} & 1.6   & 14.9  & 17.5  & 5.7   & 6.0     & 17.4  & 11.9  & 14.9  & 19.0    & 12.9  & 11.1  & 12.9  & 16.0    & 13.0 \\
    \bottomrule
    \end{tabular}%
    }
  \label{answers}%
\end{table*}%
\begin{table*}[htbp]
  \centering
  \caption{The MRR results (\%) of the model variants of Pathformer in answering EPFO queries on the BetaE datasets. }
    \begin{tabular}{cccccccccccc}
    \toprule
    \textbf{Dataset} & \textbf{Model} & \textbf{1p} & \textbf{2p} & \textbf{3p} & \textbf{2i} & \textbf{3i} & \textbf{ip} & \textbf{pi} & \textbf{2u} & \textbf{up} & \textbf{avg} \\
    \midrule
    \multirow{2}[2]{*}{\textbf{FB15k-237}} & \textbf{Pathformer-Mixer} & 44.6  & 12.9  & 10.8  & 34.3  & 48.2  & 17.1  & 26.3  & 14.6  & 10.0    & 24.3 \\
          & \textbf{Pathformer-MLP2vector} & 44.8  & 12.9  & 10.9  & 34.0    & 47.3  & 17.0    & 26.1  & 14.9  & 10.3  & 24.3 \\
    \midrule
    \multirow{2}[2]{*}{\textbf{NELL995}} & \textbf{Pathformer-Mixer} & 56.1  & 17.2  & 14.9  & 39.8  & 50.3  & 19.4  & 25.7  & 14.7  & 11.0    & 27.7 \\
          & \textbf{Pathformer-MLP2vector} & 56.6  & 17.5  & 14.9  & 39.5  & 50.1  & 19.4  & 25.8  & 14.5  & 11.2  & 27.7 \\
    \bottomrule
    \end{tabular}%
  \label{variants details}%
\end{table*}%
\begin{table*}[htbp]
  \centering
  \caption{The MRR results (\%) of experiments on Pathformer combined with other neural QE methods. The number in parentheses represents the number of layers of the path query encoder. }
  \resizebox{0.8\linewidth}{!}{
    \begin{tabular}{cccccccccccc}
    \toprule
    \textbf{Dataset} & \textbf{Model} & \textbf{1p} & \textbf{2p} & \textbf{3p} & \textbf{2i} & \textbf{3i} & \textbf{ip} & \textbf{pi} & \textbf{2u} & \textbf{up} & \textbf{avg} \\
    \midrule
    \multirow{4}[4]{*}{\textbf{FB15k-237}} & \textbf{Q2B} & 40.6  & 9.4   & 6.8   & 29.5  & 42.3  & 12.6  & 21.2  & 11.3  & 7.6   & 20.1 \\
          & \textbf{Pathformer(1)+Box embedding} & 40.0    & 10.3  & 9.1   & 29.9  & 44.6  & 9.0     & 22.8  & 10.7  & 8.8   & 20.6 \\
\cmidrule{2-12}          & \textbf{BetaE} & 39.0    & 10.9  & 10.0    & 28.8  & 42.5  & 12.6  & 22.4  & 12.4  & 9.7   & 20.9 \\
          & \textbf{Pathformer(1)+Beta embedding} & 41.2  & 11.4  & 10.1  & 29.4  & 43.3  & 13.1  & 22.9  & 12.2  & 9.7   & 21.5 \\
    \midrule
    \multirow{4}[4]{*}{\textbf{NELL995}} & \textbf{Q2B} & 42.2  & 14.0    & 11.2  & 33.3  & 44.5  & 16.8  & 22.4  & 11.3  & 10.3  & 22.9 \\
          & \textbf{Pathformer(1)+Box embedding} & 54.4      & 12.6      & 11.7      & 38.8      & 50.9      & 13.1      & 23.3      & 12.3      & 7.6      & 25.0 \\
\cmidrule{2-12}          & \textbf{BetaE} & 53.0    & 13.0    & 11.4  & 37.6  & 47.5  & 14.3  & 24.1  & 12.2  & 8.5   & 24.6 \\
          & \textbf{Pathformer(1)+Beta embedding} & 53.2      & 12.9      & 12.0      & 36.9      & 48.5      & 13.1      & 23.2      & 11.7      & 7.6      & 24.4 \\
    \bottomrule
    \end{tabular}%
    }
  \label{pathformerBoxBeta_detail}%
\end{table*}%
\begin{table*}[htbp]
  \centering
  \caption{The MRR results (\%) on FB15k-237 for Pathformer with the different number of transformer encoder layers. The number in parentheses represents the number of layers of the transformer encoder. The parameters in the table are in millions. }
  \resizebox{0.8\linewidth}{!}{
    \begin{tabular}{cccccccccccc}
    \toprule
    \textbf{Model} & \textbf{1p} & \textbf{2p} & \textbf{3p} & \textbf{2i} & \textbf{3i} & \textbf{ip} & \textbf{pi} & \textbf{2u} & \textbf{up} & \textbf{avg} & \textbf{Parameters} \\
    \midrule
    \textbf{MLP} & 42.7  & 12.4  & 10.6  & 31.7  & 43.9  & 14.9  & 24.2  & 13.7  & 9.7   & 22.6  & 9.0 \\
    \textbf{Pathformer(1)} & 44.1  & 12.7  & 10.5  & 32.0    & 44.8  & 15.4  & 24.8  & 14.4  & 10.1  & 23.2  & 12.2 \\
    \textbf{Pathformer(2)} & 44.5  & 13.0    & 10.6  & 32.7  & 46.2  & 15.9  & 25.7  & 14.3  & 10.0    & 23.7  & 19.8 \\
    \textbf{Pathformer(4)} & 44.8  & 12.9  & 10.6  & 34.1  & 46.7  & 16.7  & 26.2  & 14.8  & 10.0    & 24.1  & 35.2 \\
    \textbf{Pathformer(6)} & 44.8  & 12.9  & 10.6  & 34.2  & 47.3  & 17.0    & 26.2  & 14.9  & 10.0    & 24.2  & 50.6 \\
    \textbf{Pathformer(8)} & 44.7  & 13.1  & 10.8  & 34.3  & 47.6  & 17.3  & 26.5  & 14.8  & 10.2  & 24.4  & 66.0 \\
    \bottomrule
    \end{tabular}%
    }
  \label{layers_detail}%
\end{table*}%

\subsection{Symbolic Integration Methods for CLQA}
\label{ns method}

Recently, some studies incorporate symbolic information into neural methods. These symbolic integration methods can be also called neural-symbolic methods. 
Based on beam search, 
CQD-Beam \cite{arakelyan2020complex} uses a pre-trained KGE model \cite{trouillon2016complex} to solve complex queries. 
$\text{CQD}^{\mathcal{A}}$ \cite{arakelyan2023adapting} proposes to re-calibrate neural link prediction scores to further improve CQD-Beam. 
GNN-QE \cite{zhu2022neural} uses fuzzy logic to model logical operators and introduces a GNN \cite{zhu2021neural} to perform set projection operation. 
ENeSy \cite{xu2022neural} alleviates the cascading error problem by making neural and symbolic reasoning augment each other. 
QTO \cite{bai2023answering} is also a search-based method and considers almost all entities at every step of the search process to maximize the assignment possibility of subqueries rooted at variable nodes recursively. 
Both Pathformer and QTO focus on the tree-like query and treat the computation graph as a tree, but they process the tree differently. 
QTO starts from the leaf nodes and uses t-norm fuzzy logic and pre-trained KGE link predictor \cite{trouillon2016complex} link by link to search for the optimal solution of the variable node, 
while Pathformer uses the transformer and neural network to encode complex queries path by path. 
In addition, Pathformer is a neural method that does not use symbolic information, while the additional symbolic information introduced by QTO makes its worst-case reasoning efficiency quadratic to the size of KG.

Neural-symbolic methods not only use the training queries to train but also refer back to the training knowledge graph to obtain the symbolic information, while the query encoder of most neural methods is only learned from the queries of the training set. 
In addition, the size of the intermediate states associated with symbolic reasoning in neural-symbolic methods increases with the number of entity sets, 
whereas neural methods always operate in a fixed-size embedding space. 
Consequently, neural-symbolic methods require more computing resources than neural methods. 
In the scenario of large-scale knowledge graphs in particular, 
neural-symbolic methods are more likely to suffer from scalability problems \cite{ren2023neural, wang2023logical}. 
For example, GNN-QE employs NBFNet \cite{zhu2021neural} to compute message passing on the whole KG, resulting in complexity that is linear to $(|\mathcal{E}|+|\mathcal{V}|)d$, where $|\mathcal{E}|$ is the number of edges in KG, $|\mathcal{V}|$ is the number of nodes in KG, and $d$ is the embedding dimension. 
For Pathformer, which operates only on the query computation tree, 
the complexity is just $O(d)$. 
Besides, GNN-QE estimates the probability of whether each entity is the answer at each intermediate step, making the size of its fuzzy sets scale linearly with $|\mathcal{V}|$ \cite{zhang2024conditional}. 
As a result, GNN-QE requires much more computational costs, requiring 128GB GPU memory to run a batch size of 32. 
For Pathformer, which has only 50.6M trainable parameters, 
it only requires less than 8GB GPU memory to run a batch size of 1024. 
Combined with the above analyses, 
we consider research in this area of neural-symbolic methods to be orthogonal to the neural query embedding methods we discuss, so we do not make a direct comparison with them.

\subsection{Computational Costs}
\label{costsss}

We evaluate the computational cost of Pathformer in terms of time. 
Specifically, we compare the training speeds of Pathformer, MLP, BetaE, and GammaE. 
For a fair comparison, we use the Pathformer with a one-layer transformer encoder to conduct experiments and make the number of parameters of each method as close as possible. 
To evaluate the training speed, we calculated the average time per 100 training steps. We run them on an NVIDIA 1080 Ti. 
The experimental results are shown in Table \ref{computationcost}. 
From the experimental results, Pathformer requires more training time than MLP and less training time than others. 
Considering the performance improvement brought by Pathformer, we think the computational cost is acceptable.

\begin{table}[htbp]
  \centering
  \caption{Computational costs of Pathformer and baselines.}
    \begin{tabular}{ccc}
    \toprule
    \multirow{2}[4]{*}{\textbf{Model}} & \multicolumn{2}{c}{\textbf{Running Time per 100 steps}} \\
\cmidrule{2-3}          & \textbf{EPFO Queries} & \textbf{FOL Queries} \\
    \midrule
    \textbf{BetaE} & 85s   & 184s \\
    \textbf{MLP} & 20s   & 60s \\
    \textbf{GammaE} & 34s      & 92s \\
    \textbf{Pathformer} & 28s      & 76s \\
    \bottomrule
    \end{tabular}%
  \label{computationcost}
\end{table}%

\subsection{Statement}

This work has been submitted to the IEEE for possible publication. Copyright may be transferred without notice, after which this version may no longer be accessible.

\end{sloppypar}
\end{document}